%% file: emnlp2020.tex
\documentclass[11pt,a4paper]{article}
\usepackage[hyperref]{emnlp2020}
\usepackage{times}
\usepackage{latexsym}
\usepackage{graphicx}
\usepackage{amsmath}
\usepackage{multirow}
\usepackage{caption}
\usepackage{subcaption}
\usepackage{diagbox}
\usepackage{amsfonts}
\usepackage[inline]{enumitem}
\usepackage{booktabs,dcolumn}
\usepackage{comment}
\usepackage{xspace}
\usepackage{stfloats}
\definecolor{forestgreen}{rgb}{0.13, 0.55, 0.13}

\usepackage{microtype}

\aclfinalcopy 

\title{Fine-Grained Grounding for Multimodal Speech Recognition}

\author{Tejas Srinivasan \\
  Language Technologies Institute \\
  Carnegie Mellon University \\
  \texttt{tsriniva@andrew.cmu.edu}\\\And
  Ramon Sanabria \\
  CSTR, ILCC \\
  University of Edinburgh \\
  \texttt{r.sanabria@ed.ac.uk}\\\AND
  Florian Metze\\
  Language Technologies Institute \\
  Carnegie Mellon University \\
  \texttt{fmetze@andrew.cmu.edu}\\\And
  Desmond Elliott\\
  Department of Computer Science\\
  University of Copenhagen \\
  \texttt{de@di.ku.dk}
   \\}
 
\date{}

\newcommand{\unimodal}{\textsc{unimodal}\xspace}

\newcommand{\multibase}{\textsc{mag}\xspace}
\newcommand{\multiprop}{\textsc{maop}\xspace}

\begin{document}
\maketitle

\input{emnlp2020-templates/abstract}

\section{Introduction}
\input{emnlp2020-templates/intro}

\section{Methodology}
\label{sec:methodology}
\input{emnlp2020-templates/methodology}

\section{Experimental Setup}
\input{emnlp2020-templates/experimental-setup}

\section{Results and Analysis}
\label{sec:results}
\input{emnlp2020-templates/results}

\section{Related Work}

\input{emnlp2020-templates/relatedwork}

\section{Conclusions}
\input{emnlp2020-templates/conclusions}

\section*{Acknowledgments}
\input{emnlp2020-templates/acknowledgments}

\bibliographystyle{acl_natbib}
\bibliography{anthology,emnlp2020}

\end{document}

%% file: emnlp2020-templates/abstract.tex
\begin{abstract}
Multimodal automatic speech recognition systems integrate information from images to improve speech recognition quality, by grounding the speech in the visual context.
While visual signals have been shown to be useful for recovering entities that have been masked in the audio, these models should be capable of recovering a broader range of word types.
Existing systems rely on global visual features that represent the entire image, but localizing the relevant regions of the image will make it possible to recover a larger set of words, such as adjectives and verbs. In this paper, we propose a model that uses finer-grained visual information from different parts of the image, using automatic object proposals. In experiments on the Flickr8K Audio Captions Corpus, we find that our model improves over approaches that use global visual features, that the proposals enable the model to recover entities and other related words, such as adjectives, and that improvements are due to the model's ability to localize the correct proposals.\footnote{The code is available at \url{https://github.com/tejas1995/MultimodalASR}}
\end{abstract}

%% file: emnlp2020-templates/intro.tex
Multimodal language processing is inspired by evidence that conceptual representations in humans are distributed across modality-specific systems~\cite{barsalou2003situated}. In recent years, researchers have developed deep learning models that combine visual, linguistic, and auditory modalities for a variety of multimodal tasks, such as automatic image captioning~\cite{vinyals2015show}, visual question-answering~\cite{antol2015vqa}, and image--speech retrieval~\cite{harwath2015deep}, \textit{inter-alia}.

\begin{figure}[t]
    \centering
    \includegraphics[width=0.49\textwidth]{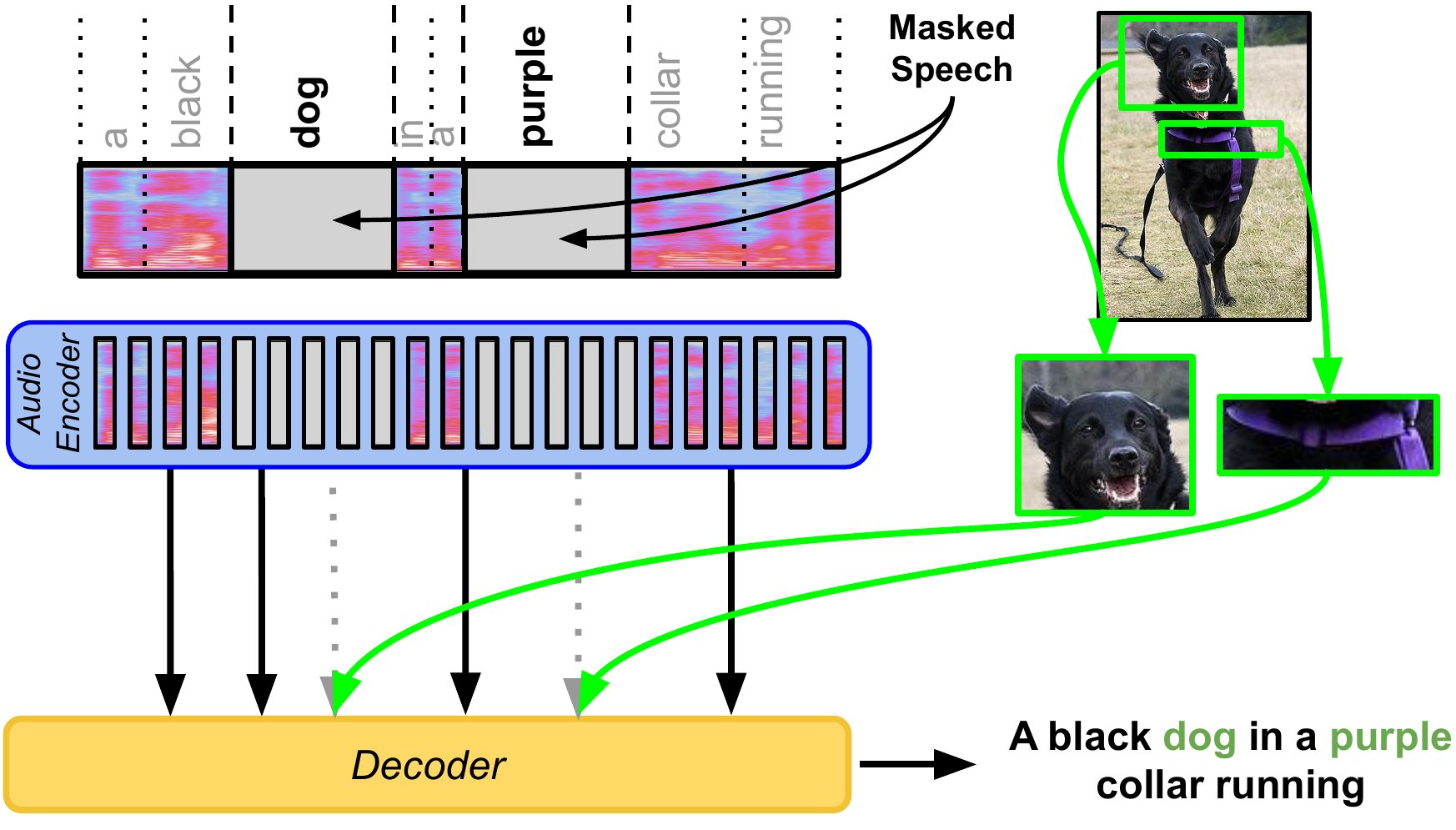}
    \caption{Our multimodal speech recognition model transcribes masked speech using visual features extracted from object proposals.}
    \label{fig:maop}
\end{figure}

In multimodal automatic speech recognition (ASR), there have been efforts to integrate visual context into acoustic models~\cite{miao2016open} and sequence-to-sequence models~\cite{palaskar2018end,sanabria18how2,Caglayan2018multimodal}. However, it is not clear if the visual context actually improves ASR or if it helps to regularize the model~\cite{Caglayan2018multimodal}. \newcite{srinivasan2020looking} recently showed that \textit{global} visual context (a single feature vector representing the entire image) is useful when the visually depictable linguistic inputs are masked,~\textit{i.e.}, masking the speech that refer to entities. This experimental methodology, inspired by~\newcite{caglayan2019probing}, creates a \textit{systematic gap} in the speech signal that can be resolved by leveraging the visual context; for example, when the audio drops during online distance-based learning or video calls with family and friends.

We present a model for multimodal ASR that learns to integrate visual features from object proposals~\cite{renNIPS15fasterrcnn}, rather than image-level features, which has previously proven to be useful for image captioning and VQA~\cite{anderson2018bottom}. Object proposals are rectangular image regions that are expected to contain objects. The novelty of our model is that when it encounters masked audio, it grounds~\cite{harnad1990symbol} the missing speech to different regions of the image. Our model learns separate attention distributions~\citep{bahdanau2016end} for each modality and combines them using a hierarchical attention mechanism in the decoder~\citep{libovicky2017attention}. This approach to integrating visual context from object proposals allows the model to better learn the relationship between speech and depicted colours, entities, and (to some extent) cardinals.

In experiments on the Flickr8K Audio Captions corpus~\citep{harwath2015deep}, we find that our model is much better at recovering masked speech than previous work. We also find that our model is right for the right reasons. In Section~\ref{sec:analysis:attention}, we perform an object localization analysis, finding that 44\% -- 49\% of the maximally attended object proposals, and 80\% -- 83\% of the top-5 attended proposals, overlap with the ground-truth bounding box annotations. This shows that our model is verifiably leveraging the visual context.

The main contributions of this paper are:
\begin{itemize}
    \itemsep0em 
    \item A new model for multimodal ASR that integrates visual features from automatically detected object proposals~\cite{renNIPS15fasterrcnn}, which make it possible for the speech to be directly grounded into regions of the image.
    \item We propose a method for forcing the model to leverage the visual context by masking a broad range of words in the speech input during training, as opposed to only masking entities~\cite{srinivasan2020looking}.
    \item We define an object localization evaluation for multimodal ASR to show when models attend to the expected regions of the image when integrating visual context.
\end{itemize}

%% file: emnlp2020-templates/methodology.tex
\subsection{Problem Formulation}

ASR is the task of transcribing a speech sequence $\mathbf{x_{1...S}}$ into a sequence of words $\mathbf{y_{1...T}}$, where $\mathbf{S}$ and $\mathbf{T}$ are the lengths of the speech and word sequence, respectively. In multimodal ASR, there is an additional visual context $\mathbf{v}$, which can be used to improve the speech transcription. In this paper, the visual context is given by a static natural image and is literally described by the speech sequence.

We investigate the utility of the additional visual context in \textit{noisy scenarios}, where words are randomly masked in the speech sequence. We expect that when the audio is clean, the audio context should be sufficient for transcription. However, when segments of the audio signal are masked, a multimodal ASR model will use the visual context to recover the missing word(s) in the speech.

\subsection{ASR Models}
\paragraph{Unimodal ASR}

Our \unimodal model is a word-level~\cite{palaskar2018acoustic} sequence-to-sequence model with attention~\cite{bahdanau2016end, chan2016listen}. The model takes as input a sequence $\mathbf{x_{1...S}}$ (as described in Section~\ref{sec:acoustic-ftrs}) which is passed through the encoder. The encoder consists of 6 bidirectional LSTM layers \cite{schuster1997bidirectional,hochreiter1997long} with temporal sub-sampling~\cite{chan2016listen} in the middle two layers. The decoder is a two-layer conditional GRU~\cite{cho2014learning} which computes attention over the encoder states $\mathbf{E}$.
\begin{align}
    \mathbf{E} & = \text{Encoder}(\mathbf{x_{1...S}}) \\
    \mathbf{h_t^{dec1}} & = \text{GRU$_1$}(\mathbf{y_{t-1}}, \mathbf{h_{t-1}^{dec1}}) \\
    \mathbf{z_t} & = \text{Attention}(\mathbf{E}, \mathbf{h_t^{dec1}}) \label{eqn:zt}\\
    \mathbf{h_t^{dec2}} & = \text{GRU$_2$}(\mathbf{z_t}, \mathbf{h_{t-1}^{dec2}}) \label{eqn:dec2}
\end{align}

\paragraph{Multimodal ASR with Global Visual Features}

The baseline multimodal ASR model uses global visual features $\mathbf{v}$ extracted from the entire image, which are incorporated into the ASR decoder. We add a hierarchical attention layer~\cite{libovicky2017attention} that adaptively weights the features from the speech encoder context vector $\mathbf{z_t}$ (Eqn. \ref{eqn:zt}) and the visual feature vector $\mathbf{v}$. The hierarchical context vector $\mathbf{z_t^{hier}}$ is the input to the second layer of the ASR decoder (Eqn. \ref{eqn:dec2}):
    \begin{align}
        \mathbf{z_t^{hier}} & = \text{Attention}(\mathbf{\{z_t, v\}}, \mathbf{h_t^{dec1}}) \\
        \mathbf{h_t^{dec2}} & = \text{GRU$_2$}(\mathbf{z_t^{hier}}, \mathbf{h_{t-1}^{dec2}})
    \end{align}
By conditioning the hierarchical attention on the output of the first decoder layer, it learns modality-specific attention weights $\alpha_a$ and $\alpha_v$ that form a probability distribution. $\alpha_a$ and $\alpha_v$ effectively control the importance of the audio and visual modalities for decoding at a given timestep. We expect that when the audio is clean, $\alpha_a$ will be higher, since clean audio is usually sufficient to transcribe a word. When the audio signal is masked, however, we expect that $\alpha_v$ will increase if the model effectively uses the visual context in the absence of information from the audio signal. We refer to this model as Multimodal ASR with Global Features (\multibase), because it utilizes global visual features.

\begin{figure*}[t]
    \centering
    \includegraphics[width=0.67\textwidth]{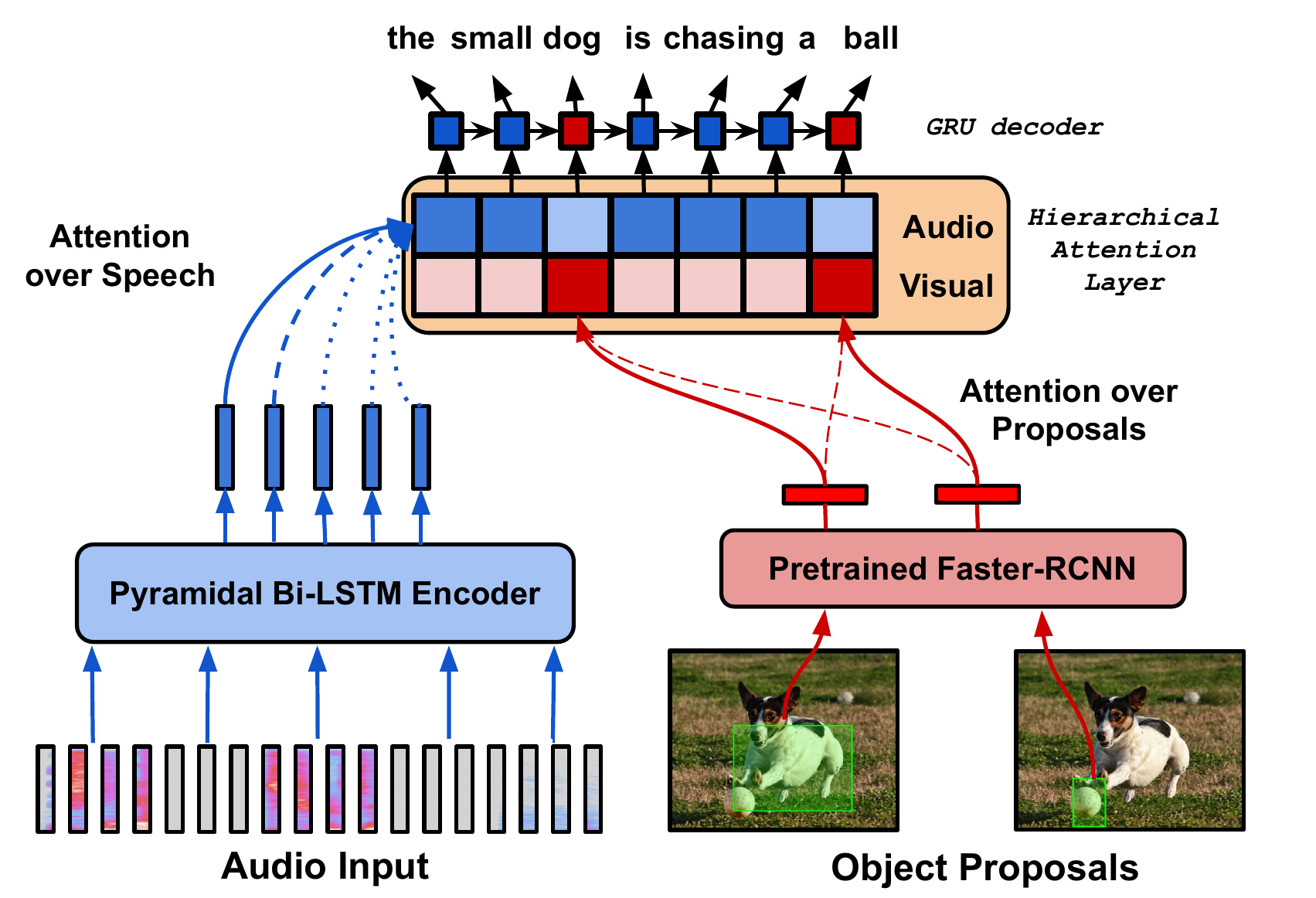}
    \caption{Multimodal ASR with Object Proposals combines attention over object proposals with attention over the audio encoding in hierarchical attention layer to correctly recover masked words in the audio input.}
    \label{fig:hierattn-prop}
\end{figure*}

\paragraph{Multimodal ASR with Object Proposals}
 Our proposed model, Multimodal ASR with Object Proposals (\multiprop), utilizes visual features from a set of object proposals, instead of the full image. The intuition behind this is that looking at object proposals can help the model localize the most important visual information at a given timestep. Identifying the relevant object proposal(s), rather than looking at the complete image, can ease the burden of transcription on the decoder. For example, it is easier for the decoder to generate a color adjective to describe an object if it extracts visual features directly from the relevant object proposal, rather than from a global visual feature vector.
 
 Concretely, for every image $\mathbf{I}$, we extract $\mathbf{N}$ object proposals $\mathbf{p_{1...N}}$, where each object proposal is a rectangular patch of the image that expected to contain an object. For each object proposal $\mathbf{p_{j}}$, we extract visual features $\mathbf{v_j}$ for that patch, in an identical manner to how they were extracted for the entire image. At every decoding timestep, the model estimates an attention distribution over the object proposal features $\mathbf{v_{1...N}}$, which gives a weighted visual representation vector $\mathbf{v^{att}_t}$. Finally, the decoder has a hierarchical attention mechanism that attends over the encoder context $\mathbf{z_t}$ and the visual representation $\mathbf{v_t^{att}}$.
 \begin{align}
        \mathbf{v_t^{att}} & = \text{Attention}(\mathbf{v_{1...N}}, \mathbf{h_t^{dec1}}) \\
        \mathbf{z_t^{hier}} & = \text{Attention}(\mathbf{\{z_t, v_t^{att}\}}, \mathbf{h_t^{dec1}}) \\
        \mathbf{h_t^{dec2}} & = \text{GRU$_2$}(\mathbf{z_t^{hier}}, \mathbf{h_{t-1}^{dec2}})
    \end{align}
 
 We want $\mathbf{v_t^{att}}$ to be representative of the most important object proposal(s) at that decoding timestep. This hierarchical attention allows the model to both identify which parts of the visual and speech context are relevant for the current decoding timestep, as well as which modality is more important. Figure~\ref{fig:hierattn-prop} illustrates the structure of our \multiprop model.

\subsection{Audio Masking}
\label{subsec:randwordmask}

Previous work has shown that the audio signal needs to be degraded during training in order to utilize the visual context~\cite{srinivasan2019analyzing}. We simulate a degradation of the audio signal during training by randomly masking words with silence. This approach extends~\newcite{srinivasan2020looking}, where they masked a fixed set of words corresponding to entities, \textit{i.e.}, objects and places. The justification for random word masking, as opposed to entity masking, is that noise in audio signals is unlikely to systematically occur when someone is speaking about an entity. Instead, multimodal ASR models should be responsive to missing audio across the linguistic spectrum.

In real-world settings, the rate at which the speech is dropped is highly variable. Therefore, we train models with an augmented version of the dataset: for each audio utterance, we create four masked audio samples, where words are masked with 0\%, 20\%, 40\% and 60\% probability. Note that the text transcript ($\mathbf{y_{1 \ldots T}}$) and image ($\mathbf{v}$) remain intact. This approach to augmenting the dataset will result in models that can adapt to different amounts of corruption in the audio signal during evaluation.

%% file: emnlp2020-templates/experimental-setup.tex
\subsection{Dataset}

We perform experiments on the Flickr 8K Audio Caption Corpus~\cite[FACC]{harwath2015deep},
which contains 40K spoken captions (total 65 hours of speech)
corresponding to 8K natural images from the Flickr8K
dataset~\cite{DBLP:conf/ijcai/HodoshYH15}. The augmented dataset that we use for training and testing (Section~\ref{subsec:randwordmask}) consists of 160K spoken captions: each caption in the original dataset has four corresponding captions in the augmented dataset.

In addition to the FACC dataset, we use the SpeechCOCO dataset~\cite{Havard2017} to pre-train our models. SpeechCOCO contains over 600 hours of synthesised speech paired with images, as opposed to natural speech in the FACC dataset.

\subsection{Acoustic Features}
\label{sec:acoustic-ftrs}

We extract 43-dimensional filter bank features from 16kHz raw speech signals. In order to mask the audio, we first extract word-audio alignments from a pre-trained Gaussian Mixture model-HMM model trained on the Wall-Street Journal Corpus, and expand the start and end timing marks by 25\% of the segment duration to account for misalignments. We mask words in the audio by replacing word segments with 0.5 seconds of silence.

\subsection{Global Visual Features}

\multibase uses a single ``global'' feature vector extracted from each image. We extract visual features from ResNet-50 CNN~\cite{DBLP:conf/cvpr/HeZRS16} pre-trained on ImageNet. We extract 2048-dim average-pooled features, and project these to 256-dim through a learned linear layer: $\mathbf{v} = \mathbf{W}\cdot\text{CNN}(\mathbf{img})$

\subsection{Object Proposal Features}
\label{sec:boundingboxes}
\multiprop uses multiple image features extracted from object proposals. We extract object proposals using a Faster-RCNN object detection model~\cite{renNIPS15fasterrcnn} with a ResNet-101 CNN backbone~\cite{DBLP:conf/cvpr/HeZRS16}. We use an implementation\footnote{\url{https://github.com/peteanderson80/bottom-up-attention}} that is pre-trained on Visual Genome dataset~\cite{krishna2017visual}. We extract a feature vector for each proposal $\mathbf{p_j}$ from the 2048-dim average pooling layer of the CNN for $\mathbf{N} = 36$ proposals. Similar to the Global Visual Features, features for each proposal are projected to 256-dim through a learned linear layer: $\mathbf{v_j} = \mathbf{W}\cdot\text{CNN}(\mathbf{p_j})$.

\subsection{Model Implementation}

All models are trained using Adam optimizer \cite{kingma2014adam}, with a learning rate of 0.0004, decay of 0.5 and batch size of 36. The encoder and decoder GRU both have 256 hidden units. The embedding dimension for the decoder is also 256, and the input and output decoder embeddings are tied \cite{press2017using}. The norm of the gradient is clipped with a threshold of 1 \cite{pascanu2012understanding}. \unimodal has 8.3M parameters, while \multibase and \multiprop have 9.1M parameters each.

Models are trained using the \textit{nmtpytorch} framework~\cite{DBLP:journals/pbml/CaglayanGBABB17}. We first pre-train our models on the SpeechCOCO dataset, which is also Augmented with masked speech. For every model described in Section~\ref{sec:methodology}, we train models on FACC using several checkpoints from the SpeechCOCO pre-training, and choose the model with the best development WER on the Augmented development set. This pre-training step, inspired by~\newcite{ilharco2019large}, was crucial to ensure stable training of our models on the FACC dataset. Models take $\approx$~ 5-6 hours to train on the FACC dataset.

\input{emnlp2020-templates/table-results}

\subsection{Evaluation Metrics}

Our model development (and the associated results) is conducted on the development set of the Flickr8K Audio Captions Corpus; the rest of our analysis is conducted on the test set. We report \textbf{Word Error Rate} (WER) for all our models, and for datasets with masked audio, we compute \textbf{Recovery Rate} (RR)~\cite{srinivasan2020looking}, which measures the percentage of masked words in the dataset that are correctly recovered in the transcription:
\begin{align*}
    \text{RR} = \frac{|\text{correctly transcribed masked words}|}{|\text{masked words in dataset}|}
\end{align*}

In addition, we calculate the contribution of the visual signal when decoding each word in the Multimodal ASR models by inspecting the attention weights of the audio and visual modalities in the hierarchical attention layer. We introduce a new metric to quantify this: \textbf{Grounding Rate}. Grounding Rate measures the percentage of correctly recovered words which had a higher visual attention weight than normal (quantified by $\mathbb{E}[\alpha_v]$). $\mathbb{E}[\alpha_v]$ is computed as the average of $\alpha_v$ over all decoding timesteps in the Augmented development set:

\begin{align*}
    \text{GR} = \frac{|\text{recovered words where } \alpha_v > \mathbb{E}[\alpha_v]|}{| \text{correctly recovered masked words}|}
\end{align*}

It has been noted that attention does not always provide a perfect explanation for an observed phenomenon ~\cite{jain2019attention, serrano2019attention}. In this paper, we examine attention to determine whether the weights align with our intuition of how the masked words are recovered, i.e. does the model recover words using the visual modality and the correct object proposal? We also use the attention distribution to conduct a quantitative object localization analysis in Section ~\ref{sec:analysis:attention}.

%% file: emnlp2020-templates/table-results.tex
\begin{table*}[t]
\centering
\begin{subtable}{\textwidth}
\centering
\begin{tabular}{lccccccccccc}
\toprule
\multicolumn{1}{c}{} & \phantom{a} & \multicolumn{4}{c}{ $\uparrow$ Recovery Rate (\%)} & \phantom{a} & \multicolumn{5}{c}{ $\downarrow$  Word Error Rate (\%)}\\
\cmidrule(lr){3-6} \cmidrule(lr){8-12}

 Masking Percentage & \phantom{a} & Aug. & 20\% & 40\% & 60\% & \phantom{a} & Aug. & 0\% (Clean) & 20\% & 40\% & 60\% \\
\midrule
\unimodal & \phantom{a} & 29.2& 37.4 & 31.4 & 25.0 \phantom{a} & & 33.8 & \textbf{13.6}  & \textbf{25.9}   & 40.2 & 56.8 \\
\midrule
\multibase     & \phantom{a} & 33.5& 40.1 & 34.9 & 30.4 \phantom{a} & & 33.3 & 13.8  & 26.1   & 39.8 & 54.8 \\
\multiprop & \phantom{a} & \textbf{36.3} & \textbf{41.5} & \textbf{37.3} & \textbf{33.2} \phantom{a} & & \textbf{32.8} & 14.1  & 26.1   & \textbf{39.1} & \textbf{53.6} \\ 
\bottomrule
\end{tabular}
\caption{Recovery Rate (RR) and Word Error Rate (WER) of the ASR models on the FACC development set.}
\label{tab:main-results}
\end{subtable}
\begin{subtable}{\textwidth}
\centering
\vspace{1em}
\begin{tabular}{ccccccccc}
\toprule

 & & Nouns   & Places  & Adjectives & Colors  & Verbs   & Adverbs & Cardinals \\ 
\cmidrule(lr){3-9}
\multirow{3}{*}{RR(\%)} & \unimodal & 37.6 & 29.3 & 27.4  & 27.2 & 27.6 & 28.4 & 56.1   \\ 
& \multibase  & 48.2 & 39.5 & 30.1  & 29.9 & \textbf{29.3} & \textbf{30.6} & 56.7   \\ 
& \multiprop  & \textbf{52.4} & \textbf{42.4} & \textbf{38.0}  & \textbf{46.0} & 29.0 & 26.7 & \textbf{57.4}   \\  
\midrule
\multirow{2}{*}{GR (\%)} & \multibase & 88.1 & 88.1 & 68.5 & 66.3 & 50.2 & 25.5 & \textbf{90.9}   \\ 
& \multiprop & \textbf{91.5} & \textbf{91.8} & \textbf{87.0} & \textbf{92.0} & \textbf{58.7} & \textbf{28.5} & 87.9 \\ \bottomrule
\end{tabular}
\caption{Comparison of Recovery Rate (RR) and Grounding Rate (GR) of our ASR models on different word categories.}
\label{tab:hierattn-results}
\end{subtable}
\caption{Results on the Flickr8K Audio Captions development set.}
\end{table*}

\normalsize

%% file: emnlp2020-templates/results.tex
In Table \ref{tab:main-results}, we summarize the performance of our three ASR models - \unimodal, \multibase and \multiprop. We examine performance on the Augmented development set, which is constructed similarly to our training set described in Section \ref{subsec:randwordmask}, consisting of samples with 0\%, 20\%, 40\% and 60\% of words masked. We also evaluate the models on datasets constructed at each individual masking level (i.e. individual datasets where words are masked with 20\%, 40\%, 60\% probability).

First, we find that the multimodal ASR models outperform the \unimodal model in terms of recovery rate, and that the difference increases as the masking rate increases from 20\% to 60\%. The Word Error Rate of the \unimodal model is slightly lower than the multimodal models for clean data, but these models perform much better than \unimodal with higher speech masking rates. Furthermore, the \multiprop model that operates over object proposals substantially outperforms the \multibase model, which uses a single global visual vector, on both metrics and at all masking levels.

We now turn our focus to analysing which types of words are best recovered by our multimodal models. We conduct this analysis across seven categories: five syntactic (nouns, verbs, adjectives, adverbs and cardinals) and two semantic (places and colors).\footnote{Syntactic categories' words were found by POS tagging the corpus and keeping the category's top 100 frequent words.} For each category, we create a new test set where we mask all occurrences of words belonging to that category.

In Table~\ref{tab:hierattn-results}, we report the recovery rate for our models on the different word categories. We see that \multibase and \multiprop are good at recovering entities (nouns and places) as well as their properties (adjectives and colors), but they perform similarly to \unimodal for other types. Furthermore, we see that while \multiprop outperforms \multibase on almost all word categories, the improvements on adjectives and colors are most significant. This shows that using object proposals gives the model a more fine-grained view of the entities and their attributes.

We also report the Grounding Rate of the multimodal models in Table~\ref{tab:hierattn-results}. When more groundable words are masked (\textit{i.e.}, entities and adjectives), the Grounding Rate is higher, indicating that the models recover these words by using the visual modality. We also see that \multiprop not only recovers more masked adjectives and colors, but also has a higher Grounding Rate for those categories. These results indicate that the model is using the hierarchical attention layer when it recovers groundable words.

\subsection{What Are You Looking At? Analyzing the Attention Over Object Proposals}
\label{sec:analysis:attention}

In the previous section, we showed that object proposals provide useful features for multimodal ASR models. We now turn our focus to examining whether this model is right for the right reasons.

We first investigate the attention distribution over object proposals, to determine if it is uniformly distributed over the proposals or concentrated over particular regions of interest. The object proposals are ranked for a given sample according to their attention weights, from which we compute the average proposal attention at each rank across all correctly recovered words in the Augmented development dataset. We observe that most of the proposal attention ($\approx 70\%$) is concentrated among the top 3 proposals, with 40\% going to a single proposal alone. This shows that not only does the model use the visual modality, it is also able to identify a proposal that it expects to be relevant for recovering a masked word.

Given that \multiprop focuses its attention distribution on one or few proposals at a time, we analyze how closely the attended object proposals match the words they are used to recover. We conduct this analysis using the ground-truth bounding box annotations from the Flickr30K Entities dataset\footnote{The Flickr30K dataset is a superset of the Flickr8K dataset. For every caption, Flickr30K Entities contains bounding box annotations for the phrases within the sentence.}~\cite{plummer2015flickr30k} by repurposing the Intersection over Union metric (IoU) from the object detection literature~\cite{russakovsky2015imagenet}. Specifically, we compute \textbf{IoU Precision @ K} as follows:
\begin{enumerate}
    \item For every correctly recovered word, we extract the top-K proposals at that decoding timestep.
    \item We find the bounding box annotation(s) in the Flickr30K dataset for all phrases in that sentence which contain the recovered word, ignoring words that do not have a bounding box annotation.
    \item From the top-K proposals and bounding boxes, we find the proposal-bounding box pair that has the highest Intersection over Union.
    \item We compute IoU Precision as the percentage of samples whose \texttt{Proposal-IoU} $>$ 0.5.
\end{enumerate}

\begin{figure}[t]
    \centering
    \includegraphics[width=0.8\linewidth]{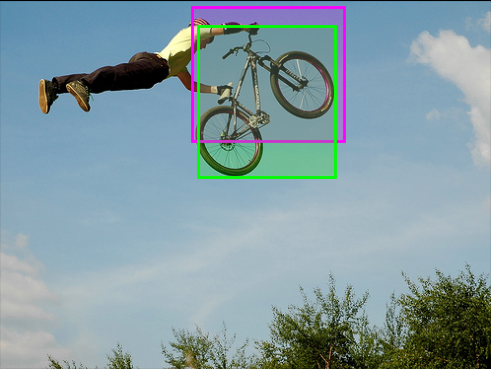}
    \caption{A \textbf{\textcolor{magenta}{localized proposal}} and \textbf{\textcolor{forestgreen}{ground truth bounding box}} for recovering 
    \textbf{bike} (IoU = 0.72)}
    \label{fig:iou-example}
\end{figure}

This metric computes the percentage of correctly recovered words for which the localized object proposal had a minimum IoU of 0.5 with a ground truth bounding box annotation. Figure~\ref{fig:iou-example} shows an example of a maximally attended proposal and a ground-truth bounding box annotation.

\input{emnlp2020-templates/table-iou}

In Table~\ref{tab:iou}, we summarize the IoU Precision @ K for our \multiprop model for different values of K, for the groundable word categories.\footnote{Verbs and adverbs did not have enough ground-truth bounding box annotations in Flickr30K Entities for this analysis. Cardinals are discussed in more detail in Section~\ref{sec:curious}.} We compare the IoU Precision from our top-K proposals with a Random-K baseline, where we pick K of the 36 object proposals randomly, instead of using the attention distribution. We see that our top-K proposals have a significantly higher IoU Precision than the Random-K baseline across all word types, with $\approx 45\%$ of the maximally attended proposals overlapping with the ground truth bounding box, and $80-83\%$ of the top-5 attended proposals overlapping. The results verify that not only is \multiprop focusing on a few proposals, but also the attended proposals are verifiably useful for recovering masked words.

\input{emnlp2020-templates/table-cleanaudio}

\subsection{Performance on Clean Speech}

\multiprop is useful for recovering words which are masked in the speech input but we also want to know how it performs on clean speech sequences. We inspect the transcriptions on a clean, unmasked version of the test set, and calculate a \textbf{Word Accuracy (WA)} for different word categories. WA captures the percentage of words belonging to the different word categories which are correctly transcribed from the clean audio signal.

In Table~\ref{tab:clean}, we observe that \multiprop performs on par with \unimodal on all word categories. This indicates that the visual modality makes no difference when the audio is clean; however, this could be an artefact of the FACC corpus, which is composed of read speech of highly structured captions, and is thus a relatively easy dataset for ASR models. We believe that in more difficult and real-world scenarios (\textit{e.g.}, with different accents and types of speech), \multiprop could use the visual modality to improve transcription without the random word masking used in this paper. 

\subsection{Case Study: The Curious Case of Cardinals}
\label{sec:curious}

\multiprop is better than \unimodal at recovering entities and their attributes but both models perform similarly at recovering masked cardinals (see Table~\ref{tab:hierattn-results}). Interestingly, the Grounding Rate of \multiprop for cardinals is high (87.9\%), which shows that the model uses the visual modality, but to limited effect. One reason for this discrepancy could be that counting entities is difficult if they are not clearly distinguishable due to visual clutter~\cite{rosenholtz2007measuring}. Another reason could be the non-uniform distribution of cardinals in the dataset: $\approx 60\%$ of the cardinals are the number \textit{two}, leading the model to learn a biased distribution.
\begin{figure}[t]
    \centering
    \includegraphics[width=\linewidth]{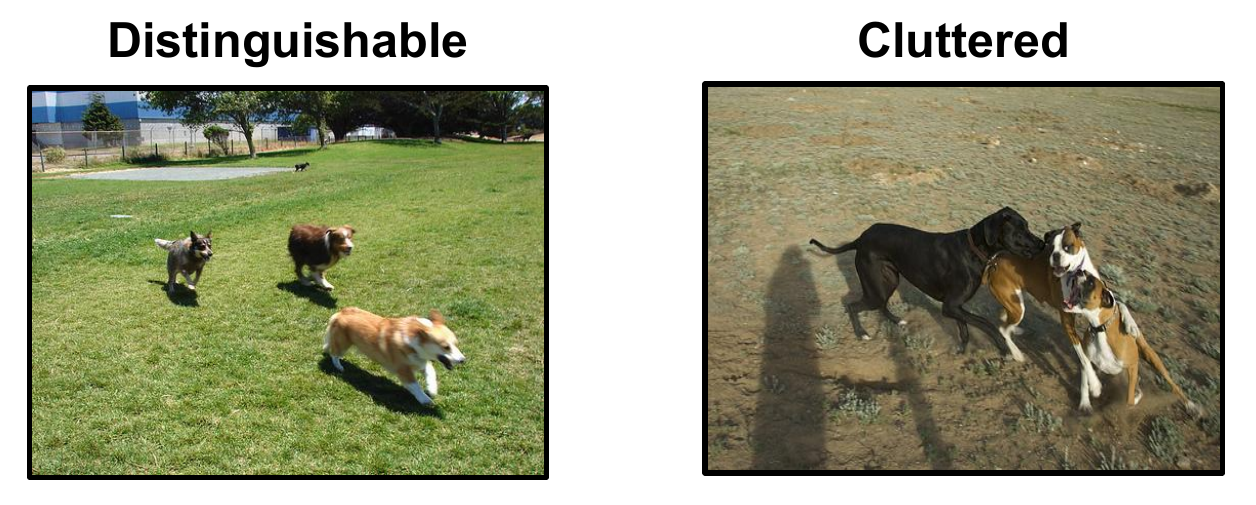}
    \caption{The images used to evaluate the ability of \multiprop to count entities (Section \ref{sec:curious}).}
    \label{fig:cardinal-swap}
\end{figure}

\input{emnlp2020-templates/table-qual}

As a case study, we evaluate the sensitivity of our model to visual clutter using 49 samples in the test set that contain the \textit{cardinal-entity} phrase ``\textit{three dogs}'' where the cardinal is masked in the audio. In these 49 samples, the model should be able to use the visual context to correctly recover the missing words ``\textit{three}'' but the recovery rate is only 24.5\%. We also chose two images from the dataset: one containing three clearly distinguishable dogs, and one containing three dogs which are hard to distinguish from each other, as shown in Figure~\ref{fig:cardinal-swap}. We proceed to calculate the recovery rate of the masked cardinal in these 49 examples with either the distinguishable or the cluttered image as the visual context, instead of the original images.

We find that recovery of \textit{three} in the noun phrase \textit{three dogs} is almost perfect using the image with the distinguishable entities (93.9\%), and very low when using the cluttered image, where the entities are hard to distinguish (2.0\%). This shows that \multiprop is capable of counting entities when they are easy to process in the visual context. Recall that the recovery rate when the original image is provided is only 24.5\%; we conducted a manual analysis of the 49 images in this case study and found that $\approx 55\%$ of them were cluttered with  entities that were hard to distinguish. We leave a more thorough analysis of a broader range of object types for future work.

\subsection{Qualitative Analysis}

Figure~\ref{tab:qual} presents qualitative examples in which words are masked in the speech sequence and recovered in the transcription. We also visualize the object proposal with maximum attention at each step, along with a relative weight of visual modality weight $\alpha_v$ in the hierarchical attention layer.

In the first example, the model correctly localizes the relevant part of the image for the two masked words (\textit{dog} and \textit{ball}) at each step and recovers these words correctly. Moreover, $\alpha_v$ is relatively higher for both those words, compared to the rest of the generated sequence. The second example is similar: the model correctly localizes the objects relevant for recovering the masked words. Interestingly, the model isolates the correct proposal for the second masked word, \textit{red}, and extracts the relevant attribute as well.

The final example shows where \multiprop both succeeds and fails. The masked words \textit{dresses} and \textit{walking} are correctly recovered using the correct locations.  However, for \textit{purple} and \textit{sign}, the model attends to the correct proposals, but fails to recover the words \textit{pink} and \textit{ship}, respectively.

We note that the object proposal attention is fairly stable across time: the same proposal is often attended to across the length of an entire phrase, rather than jumping around the image. 

%% file: emnlp2020-templates/table-iou.tex
\begin{table}[b]
\centering
\begin{tabular}{lcccc}
\toprule
Proposals & Nouns  & Places & Adj. & Colors \\ \toprule
Random-1          & 5.9  & 6.8  & 5.9      & 6.8  \\ 
Top-1        & 44.7 & 45.3 & 46.3     & 49.4 \\   
\midrule
Random-3          & 17.4 & 16.6 & 17.1     & 15.5 \\  
Top-3       & 71.7 & 68.9 & 70.2     & 71.7 \\
\midrule
Random-5          & 27.2 & 26.5 & 26.2     & 28.1 \\ 
Top-5       & 83.6 & 82.3 & 80.4     & 83.2 \\  \bottomrule
\end{tabular}
\caption{Intersection over Union Precision @ K (\%) across four different groundable word catergories.}
\label{tab:iou}
\end{table}

%% file: emnlp2020-templates/table-cleanaudio.tex
\begin{table}[t]
\centering
\begin{tabular}{cccc}
\toprule
& \unimodal & \multiprop \\
\cmidrule(lr){2-3}
Nouns      & 96.0 & 96.1 \\
Places     & 90.3 & 89.0 \\
Adjectives & 93.6 & 93.1 \\
Colors     & 94.8 & 94.2 \\
Verbs      & 93.9 & 94.1 \\
Adverbs    & 88.9 & 88.0 \\
Cardinals  & 97.1 & 97.0 \\
\bottomrule
\end{tabular}
\caption{Word Accuracy (\%) for \unimodal and \multiprop when transcribing clean, unmasked audio.}
\label{tab:clean}
\end{table}

%% file: emnlp2020-templates/table-qual.tex
\begin{figure*}
    \begin{subfigure}[b]{1\textwidth}
    \centering
    \includegraphics[width=1\linewidth]{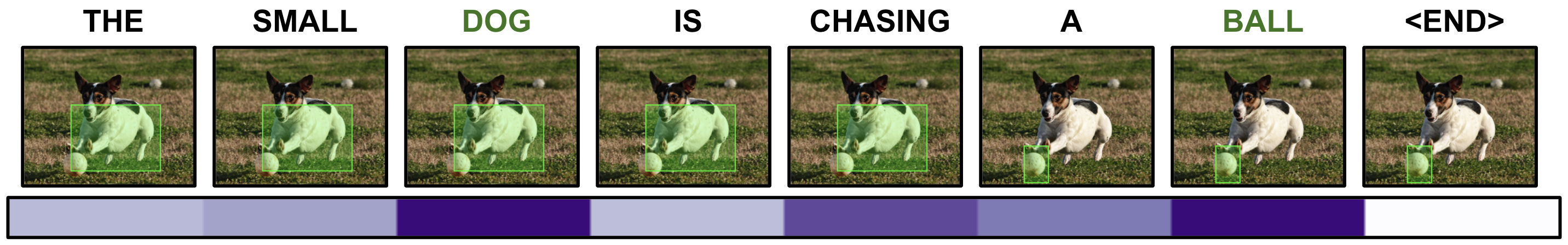}
    \end{subfigure}
    \vskip 8pt
    \begin{subfigure}[b]{1\textwidth}
    \centering
    \includegraphics[width=1\linewidth]{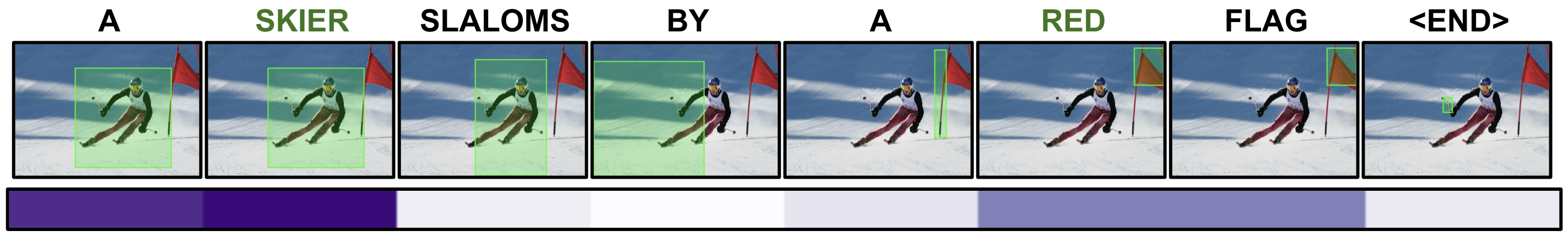}
    \end{subfigure}
    \vskip 8pt
    \begin{subfigure}[b]{1\textwidth}
    \centering
    \includegraphics[width=1\linewidth]{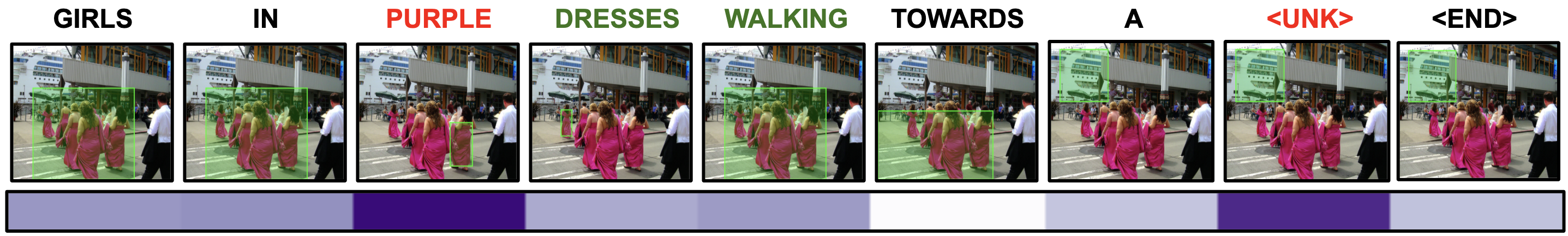}
    \end{subfigure}
\caption{Proposal and hierarchical attentions for threes samples. We present the hypothesis transcription (with \textcolor{forestgreen}{recovered} and \textcolor{red}{unrecovered} masked words), along with the maximum attended proposal (\textcolor{forestgreen}{highlighted image patch}) and \textcolor{violet}{relative hierarchical visual attention} (darker shade of purple indicates higher $\alpha_v$)  at each decoding timestep.}
\label{tab:qual}
\end{figure*}

%% file: emnlp2020-templates/relatedwork.tex
Inspired by studies of human perception, multimodal processing is spreading into many traditional areas of research, \textit{e.g.}, machine translation ~\cite{sulubacak2019multimodal} and ASR~\cite{palaskar2018end}. It has become an important part of new areas of research such as  image captioning~\cite{bernardi2016automatic}, visual question-answering (VQA; \cite{antol2015vqa}), and multimodal summarization~\cite{palaskar2019multimodal}. 

The representation and integration of visual context in multimodal ASR systems is an active area of research. Previous approaches incorporate image representations either in the acoustic model~\cite{miao2016open}, the language model~\cite{gupta2017visual,Naszadi_2018_ECCV_Workshops}, or in end-to-end models~\cite{sanabria18how2}.~\citet{Caglayan2018multimodal} and~\citet{moriya2018lstm} explore different types of multimodal representations such as image-scene representations and titles of instructional videos respectively. Although all these integration methods show improvements over unimodal baselines, it is not clear when such approaches perform better, and which representations are best.

It has been argued that traditional multimodal architectures do not necessarily take advantage of image semantics in different tasks.~\citet{Caglayan2018multimodal} showed that multimodal ASR models trained with~\textit{shift adaptation}~\cite{miao2016open}\footnote{A linear transformation conditioned on the visual features is applied on the audio features.} use the image as a regularization signal. In a similar direction, ~\citet{desmond} showed that misaligment between image and text representations do not affect multimodal MT models. \citet{ramakrishnan2018overcoming} and ~\citet{grand2019adversarial} showed that traditional VQA neural architectures ignore the visual context and focus on linguistic biases of the dataset. More related to our work are the studies of ~\citet{srinivasan2020looking} and~\newcite{caglayan2019probing}, which explore how multimodal models use image information under noisy scenarios. These studies conclude that when certain nouns are dropped from the dominant language modality, multimodal models are capable of properly using the semantics provided by the image. However, unlike this work, their explorations are limited to nouns and not expanded to other types of words.

From an image representation perspective, previous works have studied the utility of using local representations, rather than global ones for multimodal language processing tasks. For instance, ~\citet{xu2015show} show that, by using attention, the model can use different regions of the image while performing image captioning. More recent work shows that bounding boxes~\cite{renNIPS15fasterrcnn}, a discrete variant of attention over images, improve the representation and hence the performance of different tasks such as VQA~\cite{anderson2018bottom}, image captioning~\cite{yin2017obj2text} and machine translation~\cite{grounded}. In this work, we apply this methodology to multimodal ASR (see Section~\ref{sec:boundingboxes}).

%% file: emnlp2020-templates/conclusions.tex
In this work, we introduce a new model for multimodal ASR that attends overs fine-grained object proposals and is capable of recovering words which are masked in the speech signal. We show that our model recovers masked words because it can accurately identify the relevant object proposal(s), and that this ability allows it to not only recover the object when it has been masked in the speech signal, but also the object's attributes.

In future work, we plan to improve our model by masking random speech segments~\citep{park2019specaugment} rather than aligned words. If successful, this methodology would allow us to train and test our multimodal models without the need for word alignments, a current limitation of our framework. We will also experiment with more challenging speech captioning scenarios where speech ambiguities are more likely to occur~\cite{PontTuset_arxiv2019}.

%% file: emnlp2020-templates/acknowledgments.tex
This work used the computational resources of the PSC Bridges cluster at Extreme Science and Engineering Discovery Environment (XSEDE)~\cite{6866038}. We thank Stella Frank for discussions about whether such a model could be expected to count objects in images.